\documentclass[sigconf,authorversion]{acmart} 

\usepackage{amsmath}
\usepackage{amsfonts}
\usepackage{hyperref}
\usepackage{graphicx}
\usepackage{subcaption}
\usepackage{caption}
\usepackage{cprotect}
\usepackage{balance}

\AtBeginDocument{%
  \providecommand\BibTeX{{%
    \normalfont B\kern-0.5em{\scshape i\kern-0.25em b}\kern-0.8em\TeX}}}

\copyrightyear{2022}
\acmYear{2022}
\setcopyright{acmcopyright}
\acmConference[CIKM '22] {Proceedings of the 31st ACM International Conference on Information and Knowledge Management}{October 17--21, 2022}{Atlanta, GA, USA.}
\acmBooktitle{Proceedings of the 31st ACM International Conference on Information and Knowledge Management (CIKM '22), October 17--21, 2022, Atlanta, GA, USA}
\acmPrice{15.00}
\acmISBN{978-1-4503-9236-5/22/10}
\acmDOI{10.1145/3511808.3557275}

\begin{document}

\title{Cross-Domain Aspect Extraction using Transformers Augmented with Knowledge Graphs}


\author{Phillip Howard}
\email{phillip.r.howard@intel.com}
\affiliation{%
  \institution{Intel Labs}
  \city{Chandler}
  \state{Arizona}
  \country{USA}
}
\author{Arden Ma}
\affiliation{%
  \institution{Intel Labs}
  \city{Santa Clara}
  \state{California}
  \country{USA}
}
\author{Vasudev Lal}
\affiliation{%
  \institution{Intel Labs}
  \city{Hillsboro}
  \state{Oregon}
  \country{USA}
}
\author{Ana Paula Simoes}
\affiliation{%
  \institution{Intel Labs}
  \city{Santa Clara}
  \state{California}
  \country{USA}
}
\author{Daniel Korat}
\affiliation{%
  \institution{Intel Labs}
  \city{Petah Tikva}
  \country{Israel}
}
\author{Oren Pereg}
\affiliation{%
  \institution{Intel Labs}
  \city{Petah Tikva}
  \country{Israel}
}
\author{Moshe Wasserblat}
\affiliation{%
  \institution{Intel Labs}
  \city{Petah Tikva}
  \country{Israel}
}
\author{Gadi Singer}
\affiliation{%
  \institution{Intel Labs}
  \city{Santa Clara}
  \state{California}
  \country{USA}
}

\renewcommand{\shortauthors}{Phillip Howard et al.}

\begin{abstract}
The extraction of aspect terms is a critical step in fine-grained sentiment analysis of text. Existing approaches for this task have yielded impressive results when the training and testing data are from the same domain. However, these methods show a drastic decrease in performance when applied to cross-domain settings where the domain of the testing data differs from that of the training data. To address this lack of extensibility and robustness, we propose a novel approach for automatically constructing domain-specific knowledge graphs that contain information relevant to the identification of aspect terms. We introduce a methodology for injecting information from these knowledge graphs into Transformer models, including two alternative mechanisms for knowledge insertion: via query enrichment and via manipulation of attention patterns. We demonstrate state-of-the-art performance on benchmark datasets for cross-domain aspect term extraction using our approach and investigate how the amount of external knowledge available to the Transformer impacts model performance. 
\end{abstract}





\maketitle

\section{Introduction}
\label{sec:introduction}
Sentiment analysis is a fundamental task in NLP which has been widely studied in a variety of different settings. While the majority of existing research has focused on sentence- and document-level sentiment extraction, there is considerable interest in fine-grained sentiment analysis that seeks to understand sentiment at a word or phrase level. For example, in the sentence {\it ``The appetizer was delicious''}, it may be of interest to understand the author's sentiment regarding a specific aspect ({\it appetizer)} in the form of an expressed opinion ({\it delicious}). This task is commonly referred to as Aspect-Based Sentiment Analysis (ABSA). 

ABSA is often formulated as a sequence tagging problem, where the input to a model is a sequence of tokens $X = \{x_{1}, x_{2}, ..., x_{n}\}$. For each token $x_{i} \in X$, the objective is to correctly predict a label $y_{i} \in \{BA, IA, BO, IO, N\}$. The labels $BA$ and $IA$ denote the beginning and inside tokens of aspect phrases while $BO$ and $IO$ indicate the beginning and inside tokens of opinions. The class $N$ denotes tokens that are neither aspects nor opinions. The focus of our work is improving the identification of aspects within the context of the ABSA sequence tagging problem.

Existing work on aspect term extraction has achieved promising results in single-domain settings where both the training and testing data arise from the same distribution. However, such methods typically perform much worse when the training (or {\it source}) domain differs from the testing (or {\it target}) domain. This cross-domain setting for aspect extraction poses a greater challenge because there is often very little overlap between aspects used in different domains. For example, aspects prevalent in consumer reviews about laptops (e.g., {\it processor},  {\it hardware}) are unrelated to common aspects in restaurant reviews (e.g., {\it food}, {\it appetizer}).

To address this challenging task, we introduce a novel method for enhancing pretrained Transformer models \cite{transformers} with information from domain-specific knowledge graphs that are automatically constructed from semantic knowledge sources. We show how injecting information from these knowledge graphs into Transformer models improves domain transfer by providing contextual information about potential aspects in the target domain.

This work consists of four primary contributions. First, we introduce an approach for constructing domain-specific knowledge graphs from unlabeled text using an existing large-scale commonsense knowledge graph (ConceptNet, \citet{Speer_Chin_Havasi_2017-conceptnet}) and a Transformer -based generative knowledge source fine-tuned for the task of predicting relations within a domain (COMET, \citet{Bosselut2019COMETCT-comet}). Second, we present a methodology for determining when it is beneficial to inject external knowledge into a Transformer model for aspect extraction through the application of syntactic information. Third, we explore two alternative approaches for injecting knowledge into language models: via insertion of a pivot token for query enrichment and through a disentangled attention mechanism. Experimental results demonstrate how this methodology achieves state-of-the-art performance on cross-domain aspect extraction using benchmark datasets from three different domains of consumer reviews: restaurants, laptops and digital devices \cite{pontiki-etal-2014-semeval,pontiki-etal-2015-semeval,wang-etal-2016-recursive}. Finally, we contribute an improved version of the benchmark digital devices dataset to facilitate future work on aspect-based sentiment analysis.

\section{Related Work}

\subsection{Knowledge Graphs}

A variety of knowledge graphs have been created in recent years to store large quantities of factual and commonsense knowledge about the world. ConceptNet is a widely-used and freely-available source of commonsense knowledge that was constructed from both expert sources and crowdsourcing. A variety of solutions that leverage ConceptNet have been developed for NLP tasks in recent years, including multi-hop generative QA \cite{bauer2018commonsense}, story completion \cite{chen2019incorporating}, and machine reading comprehension \cite{xia2019incorporating}.

The main challenge in using ConceptNet is the selection and quality assessment of paths queried from the graph to produce relevant subgraphs for downstream use. A variety of heuristic approaches have been proposed for this task, including setting a maximum path length \cite{guu2015traversing}, limiting the length of the path based on the number of returned nodes \cite{boteanu2015solving}, and utilizing measures of similarity calculated over embeddings \cite{gardner2014incorporating}. Auxiliary models that assess the naturalness of paths have also been proposed for predicting path quality \cite{zhou2019predicting}.

\subsection{Domain Adaptation}
Developing models that can generalize well to unseen and out-of-domain examples is a fundamental challenge in robust solution design. A key objective of many previous domain adaptation approaches has been to learn domain-invariant latent features that can be used by a model for its final predictions. Prior to the widespread usage of Deep Neural Networks (DNNs) for domain adaptation tasks, various methods were proposed that attempted to learn the latent features by constructing a low-dimensional space where the distance between features from the source and target domain is minimized \cite{TransferLearningViaDimRed, DomainAdaptationViaTransCompAnalysis}.

With the recent introduction of DNNs for domain adaptation tasks, there has been a shift towards monolithic approaches in which the domain-invariant feature transformation is learned simultaneously with the task-specific classifier as part of the training process. These methods incorporate mechanisms such as a Gradient Reversal Layer \cite{UnsupervisedDomainAdaptationByBackprop} and explicit partitioning of a DNN \cite{DomainSeparationNetworks} to implicitly learn both domain-invariant and domain-specific features in an end-to-end manner.

Such approaches have been applied to various problems in NLP, including cross-domain sentiment analysis. \citet{du-etal-2020-adversarial} and \citet{gong-etal-2020-unified}
introduce additional training tasks for BERT \cite{devlin-etal-2019-bert} in an effort to learn both domain-invariant and domain-specific feature representations for sentiment analysis tasks. The utilization of syntactic information has also been shown to be an effective way of introducing domain-invariant knowledge, which can help bridge the gap between domains \cite{Ding2017crossdomain,Wang2018cross_domain,pereg-etal-2020-syntactically-libert,lal-etal-2021-interpret}.

\subsection{Knowledge Informed Architectures}

An alternative paradigm for developing robust solutions is to augment models using external knowledge queried from a large non-parametric memory store, commonly known as a Knowledge Base (KB) or Knowledge Graph (KG). We refer to this class of models as knowledge informed architectures. Much of the existing work on knowledge informed architectures augments BERT \cite{devlin-etal-2019-bert} with external knowledge from sources such as WordNet \cite{wordnet} and ConceptNet. These approaches have led to a myriad of new BERT-like models such as KnowBERT \cite{Peters2019KnowledgeEC/knowbert}, K-BERT \cite{liu2020k}, and E-BERT \cite{poerner-etal-2020-e/e-bert} which attempt to curate and inject knowledge from KBs in various ways. How knowledge is acquired and used in these models is highly task dependent.

Knowledge informed architectures have been shown to be effective at a variety of tasks, achieving superior performance in recent challenges such as Efficient Question-Answering \cite{Min2021NeurIPS2EQA} and Open-domain Question-Answering \cite{Lewis2020RetrievalAugmentedGF, Shuster2021RetrievalAR} where external knowledge is used to enrich input queries with additional context that supplements the implicit knowledge stored in the model's parameters. To the best of our knowledge, no previous knowledge informed architectures have been developed for cross-domain aspect extraction.

\section{Methodology}
\label{sec:methodology}
\begin{figure*}[h!]
     \centering
         \includegraphics[width=1\textwidth]{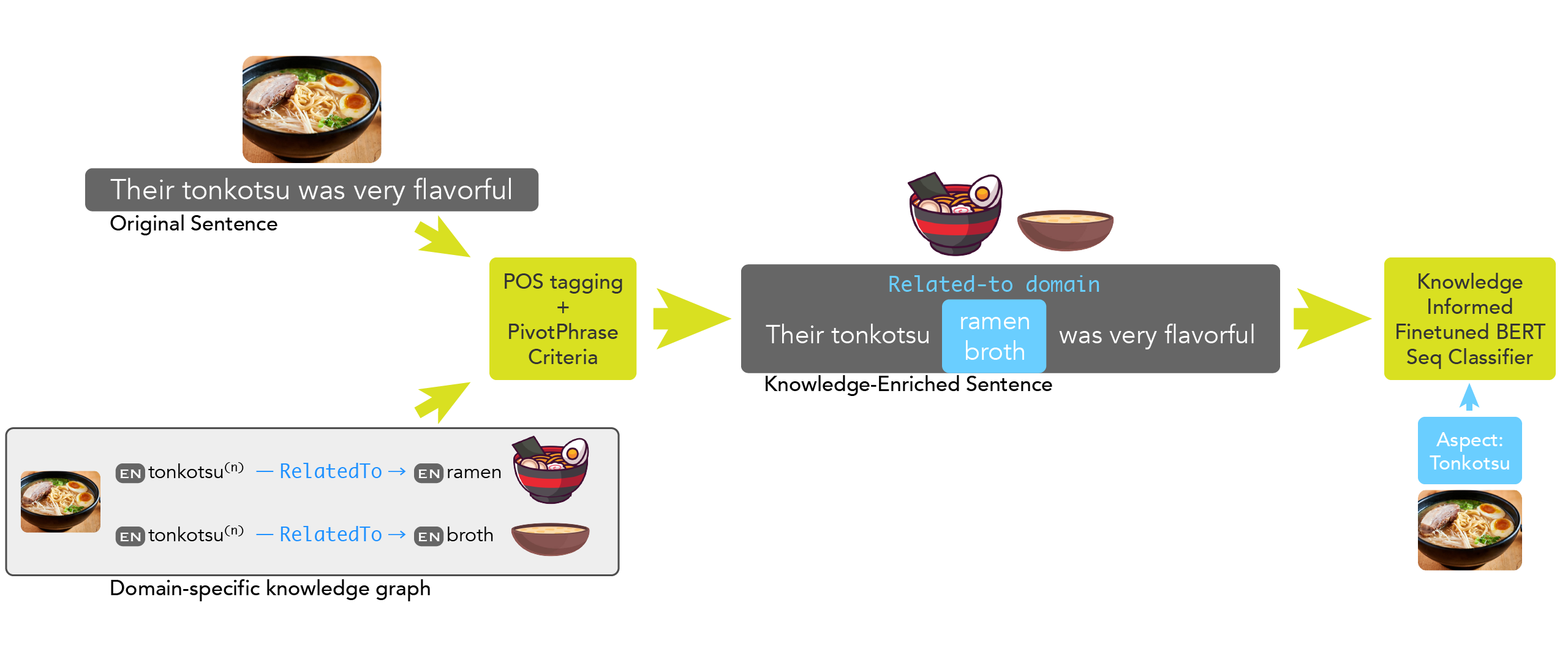}
         \caption{Illustration of our pivot token knowledge injection approach for aspect extraction}
         \label{fig:injection_diagram}
\end{figure*}

Our approach consists of a three-step process: (1) preparing a domain-specific KG for each of the target domains, (2) determining when the model can benefit from external information, and (3) injecting knowledge retrieved from the KG when applicable. We explore two alternative methods for the final knowledge injection step of this process: insertion of a pivot token into the original query, and knowledge injection into hidden state representations via a disentangled attention mechanism. We provide an illustration of our approach in Figure~\ref{fig:injection_diagram} and detail the methodology for each step of the process in the subsequent sections.

\subsection{Domain-Specific KG Preparation}

In order to ground the model in concepts related to the target domain, we create a domain-specific KG by first querying a subgraph from ConceptNet using a list of seed terms that are related to the domain. For each target domain $d$, the seed term list $S_{d} = \{s_{1}, s_{2}, ..., s_{k}\}$ is generated by applying TF-IDF to all of the unlabeled text in domain $d$ and identifying the top-$k$ scoring noun phrases. We use $k = 7$ seed terms in this work but note that the number of seed terms can be adjusted based on the desired size of the KG.

For each seed term $s \in S_{d}$, we query ConceptNet for all English-language nodes connected by an edge to $s$ and add them to the domain-specific subgraph along with the seed term $s$. The subgraph is further expanded by iteratively querying additional nodes that are connected by an edge to a node already present in the subgraph, up to a maximum distance of $h$ edges from $s$. In our experiments, we utilized a maximum edge distance of $h = 2$ for efficiency and based on the observation that querying beyond two edges from a given node in ConceptNet does not significantly increase the identification of domain-relevant concepts. 

To increase the relevancy of the queried subgraph to the target domain, we prune nodes on paths in the graph that have a low relatedness score to the seed term $s$ from which they originated. While our approach is compatible with various embedding methods, we utilize pre-computed ConceptNet Numberbatch embeddings \cite{Speer_Chin_Havasi_2017-conceptnet} that combine information from the graph structure of ConceptNet with other word embedding techniques such as Word2vec \cite{mikolov2013distributed} and GloVe \cite{pennington2014glove}. Let $\mathbf{e}_{i}$ denote the embedding vector of a given node $i$ in the subgraph. The relatedness score $r_{i,j}$ for a pair of nodes $i$ and $j$ in the graph is calculated as the cosine similarity between their embedding vectors:

\begin{equation}
    r_{i,j} = \frac{\mathbf{e}_{i} \cdot \mathbf{e}_{j}}{\| \mathbf{e}_{i} \| \| \mathbf{e}_{j} \|}
\end{equation}

For a given path $P = \{n_{s}, n_{1}, ..., n_{h}\}$ connecting nodes $n_{1} , ... n_{h}$ to the node $n_{s}$ corresponding to seed term $s$, we calculate its minimum path relatedness score, denoted $P_{\min}$, as the minimum of the pairwise relatedness scores between each node in the path and the seed term:

\begin{equation}
    P_{\min} = \min_{\forall i \in \{1, .., h\}} r_{s, i}
\end{equation}

Nodes terminating a path for which $P_{\min} < 0.2$ are discarded from the subgraph, where this threshold was chosen heuristically and can be tuned based on the application. This path filtering criteria helps disambiguate edges in ConceptNet for words that have multiple different meanings. 
Higher values of $P_{\min}$ reduce the number of unrelated nodes in the subgraph at the cost of decreased coverage.

To further expand the coverage of the domain-specific KGs, we employ a generative commonsense model called COMET \cite{Bosselut2019COMETCT-comet} to automatically augment the KG with additional terms that are related to those already present in the graph. Given a head $h$ and relation $r$, COMET is trained to predict the tail $t$ completing an $(h,r,t)$ triple. We chose to use COMET for augmenting our KGs due to the incompleteness of ConceptNet, which can vary significantly in coverage across domains as a result of its reliance on crowdsourcing for knowledge acquisition.

The original implementation of COMET consisted of GPT \cite{Radford2018ImprovingLU-gpt} fine-tuned to complete $(h,r,t)$ triples sampled from either ConceptNet or ATOMIC \cite{atomic_2019}. Motivated by the observation that the original COMET lacks coverage for certain concepts in our target domains, we improve the relevancy of its predictions by fine-tuning COMET on ConceptNet triples that are selectively chosen. For each target domain, we identify all nouns and noun phrases occurring in its text using spaCy \cite{spacy} and then query ConceptNet for triples that contain one of these nouns. A domain-specific instance of COMET is then trained by fine-tuning GPT on the task of $(h,r,t)$ completion using only the sampled set of domain-relevant triples. For each seed term $s \in S_{d}$, we use our domain-tuned implementation of COMET to generate 100 completions for the triple $(s, \text{RelatedTo}, t)$ and add them to the domain-specific KG if they are not already present.

\subsection{Determining when to Inject Knowledge}
\label{sec:whentoinject}

To determine when to inject knowledge, we identify tokens that are potential aspects by first using spaCy to extract POS and dependency relations. Motivated by the observation that aspects tend to be either individual nouns or noun phrases, we extract the candidate set of tokens by identifying noun forms in the input sequence. Multi-word phrases are extracted when one or more adjacent tokens have a dependency relation of either "amod" or "compound" and are followed immediately by a noun. Examples of multi-word phrases identified using this criteria include "iMac backup disc" and "external hard drive."

The set of tokens extracted by this process are then compared to the domain-specific KG to determine which should be flagged as related to the domain. For single-token nouns, we look for an exact match to one of the nodes appearing in the domain-specific KG. We also search for an exact match to multi-word noun phrases, but iteratively shorten the phrase by removing the left-most token if an exact match is not found. This iterative approach is used to identify the longest subset of the noun phrase that is present in the KG and is based on the observation that the right-most token in a compound noun is typically the head, which conveys the main meaning of the phrase.

This process results in the identification of a set of tokens within each sentence which are more likely to be aspects due to their syntactic context and relation to the target domain. The final step in our approach injects this knowledge into the language model to improve the accuracy of aspect classification.

\subsection{Knowledge Injection Mechanisms}
\label{sec:injectionmechanisms}
We explore two alternative methods for injecting knowledge into Transformer models. The first approach enriches the input query by inserting a pivot token after tokens identified as potential aspects as described in the previous section. The second approach is inspired by DeBERTa's \cite{he2020deberta} Disentangled Attention mechanism and utilizes the decomposition of attention to condition each token's attention distribution on the pivoting information.

\subsubsection{Knowledge injection via Pivot Token}
\label{sec:pivotphrase}

The pivot token is a special token that serves the purpose of indicating to the model that the preceding token has a greater likelihood of being labeled an aspect. We reserve two distinct pivot tokens in a model's vocabulary, one corresponding to the $BA$ class ([\textsc{domain-b}]) and another to the $IA$ class ([\textsc{domain-i}]). The [\textsc{domain-b}] pivot token is inserted after single-token aspect candidates or after the first token in a multi-word phrase, with the [\textsc{domain-i}] pivot token being used for the remaining tokens in multi-word phrases.

The following sentence is an example of a query from a restaurant review that was enriched via this process: "{\it It was the best pad} [\textsc{domain-b}] {\it thai} [\textsc{domain-i}] {\it I've ever had}." In this example, "{\it pad thai}" was marked for knowledge injection based on its syntactic information and presence in the domain-specific KG.

While the criteria described in Section~\ref{sec:whentoinject} is used to determine when to insert pivot tokens in the target domain datasets, we use a different method of stochastic pivot token insertion for the training data in order to teach the Transformer model how to use the injected knowledge. Specifically, we define hyperparameters $p$ and $r$ that correspond to the desired precision and recall of the pivot token (respectively) when used to identify aspects. We then perform a stochastic insertion of pivot tokens after a portion of the labeled aspects in the training dataset as well as some non-labeled tokens such that the precision and recall of the pivot token approximates $p$ and $r$. The purpose of this is to adapt the base language model to potential inaccuracies in the pivot token insertions while removing any dependency between the coverage of the KGs in the source and target domains.

\subsubsection{Knowledge Injection Using Disentangled Attention}
\label{sec:disentangledatt}
The second approach we consider for injecting knowledge consists of modifying the attention patterns in a Transformer model based on the candidate aspect terms identified. Inspired by the success of DeBERTa on a variety of NLU benchmarks, we utilize the Disentangled Attention mechanism introduced in DeBERTa and augment it with new attention score terms that encode positional information about the location of the candidate aspect terms. The motivation for this approach is twofold. First, it preserves the structure of the original input sequence by not requiring the injection of additional tokens. Second, it allows for finer-grained control over the attention patterns exhibited in the model.

In DeBERTa, each token $t_i$, $i = 1...N$ in the input sequence is represented by two embeddings: a content embedding $c_i$ and a position embedding $p_i$. 
This decomposed representation leads to the formulation of Disentangled Attention as follows.

\begin{equation}
\begin{split}
    A_{i,j}^{c2c} &= {Q_i^c K_j^c}^T, A_{i,j}^{c2p} = Q_i^c K_{\delta(i,j)}^{p^T }\\
    A_{i,j}^{p2c} &= K_j^c Q_{\delta(j,i)}^{p^T} \\
    A_{i,j} &= A_{i,j}^{c2c} + A_{i,j}^{c2p} + A_{i,j}^{p2c} \\
    H &= \textrm(\frac{A}{\sqrt{3d}})V^c
\end{split}
\end{equation}

Here $Q^c, K^c, V^c \in \mathbb{R}^{N \times d}$ are the Query, Key, and Value projections for the content embeddings, and $Q^p, K^p \in \mathbb{R}^{N \times d}$ are the Query and Key projections for the position embeddings. $\delta(i,j) \in [0,2k)$ is the relative distance between token $i$ and token $j$ where $k$ is the maximum relative distance possible. The joint attention matrix $A$ is then used to construct the next set of hidden states $H$ in the standard manner.

To encode the pivoting information, we define two new learned embedding vectors $m^+, m^- \in \mathbb{R}^d$ to denote whether or not a token is a candidate aspect term (respectively). We use the learned embedding vectors to create a sequence of embeddings $S^m = m_1 ... m_N$ where
\[
    m_i = 
    \begin{cases}
        m^+ \textrm{ if } t_i \textrm{ is a candidate aspect}\\
        m^- \textrm { otherwise}
    \end{cases}
\]

The Query and Key projections for these embedding vectors $Q^m, K^m \in \mathbb{R}^{N \times d}$ are learned and used in our modified attention formulation (Equation~\ref{eq:modified_dea}) through two new constituent terms, $A^{c2m}$ and $A^{m2c}$.

\begin{equation}
\label{eq:modified_dea}
\begin{split}
    A_{i,j}^{c2m} &= {Q_i^c K_{j}^m}^T,
    A_{i,j}^{m2c} = {Q_i^m K_j^c}^T \\
    \hat{A}_{i,j} &= A_{i,j}^{c2c} + A_{i,j}^{c2p} + A_{i,j}^{p2c} \\ &+ A_{i,j}^{c2m} + A_{i,j}^{m2c} \\
    \hat{H} &= \textrm(\frac{\hat{A}}{\sqrt{5d}})V^c
\end{split}
\end{equation}

Intuitively, the $A^{c2m}$ and $A^{m2c}$ terms act as a mechanism by which the model can adjust the attention distribution for a given token based on relationships between the content representations of tokens created by the Transformer and the pivoting information in $S^m$. 
We hypothesize that the $A^{c2m}$ and $A^{m2c}$ terms together encourage the model to learn attention patterns that highlight contributions from the candidate aspect terms, leading to hidden state representations that carry additional relevant information about the locations of potential aspects.

\section{Experiments}
\subsection{Experimental Setup}
\label{sec:experiments}
We evaluate the cross-domain aspect extraction performance of our approach on three benchmark ABSA datasets consisting of English-language consumer reviews for restaurants (5,841 sentences), laptops (3,845 sentences), and digital devices (3,836 sentences) \cite{pontiki-etal-2014-semeval,pontiki-etal-2015-semeval,wang-etal-2016-recursive}. This three-dataset experimental setting is one of the largest available for ABSA, which is more limited in data availability than other sentiment analysis tasks due to the difficult and time-consuming nature of labeling aspects. To assess cross-domain performance, we create pairings of the three data domains as follows: let $L$, $R$, and $D$ denote the laptops, restaurants, and device review datasets (respectively). The cross-domain settings on which we evaluate our models is represented by the set $\mathcal{D}$ in Equation~\ref{eq:dataPairings}, where the first element in each tuple is the source domain and the second element is the target domain. 
\begin{equation}
    \label{eq:dataPairings}
    \mathcal{D} = \{ (L,R), (L,D), (R,L), (R,D), (D,L), (D,R)\}
\end{equation}

Within each domain, we create 3 separate partitions of the data which are then further divided into a train, validation, and test set following a 3:1:1 ratio. The performance of each model is evaluated by first being fine-tuned for ABSA on the source domain data and then being tested on the target domain data. To control for elements of randomness, we repeat each of our experiments using three different random seeds, reporting the mean and standard deviation of the aspect extraction F1 score calculated over all combinations of random seeds and data partitions (9 in total). In accordance with prior work, only exact matches between the predicted labels and gold labels are counted as correct.

We adopt the HuggingFace \cite{wolf-etal-2020-transformers} implementations of Transformer models used in our experiments and open-source our code\footnote{Our code is available via \href{https://github.com/IntelLabs/nlp-architect/tree/aspect_extraction_with_kg}{NLP Architect}.}. Following the experimental setup of \citet{pereg-etal-2020-syntactically-libert}, we use the validation dataset to determine when to apply early stopping as well as the hyperparameters $p$ and $r$ defined in Section~\ref{sec:methodology}. Specifically, we use a heuristic approach of setting $p$ and $r$ equal to the evaluated precision and recall of the pivot token insertions on the validation dataset. Other hyperparameters were chosen by adopting the same configuration used previously by \citet{pereg-etal-2020-syntactically-libert}, which includes fine-tuning the model using the AdamW optimizer \cite{Loshchilov2019DecoupledWD} with a learning rate of $5 \times 10^{-5}$, a batch size of 8, and a maximum sequence length of 64 tokens for up to 10 epochs. 
Our experiments were conducted on a Ubuntu 18.04 system with an Intel(R) Xeon(R) Platinum 8280 CPU and three Nvidia RTX 3090 GPUs.

\subsection{Results}

\begin{table}[t]
\resizebox{0.5\textwidth}{!}{%
\centering
\begin{tabular}{p{2.1cm} p{0.75cm} p{0.75cm} p{0.75cm} p{0.75cm} p{0.75cm} p{0.75cm} p{0.75cm}}
\hline
\textbf{Model} & \textbf{$(L,R)$} & \textbf{$(L,D)$} & \textbf{$(R,L)$} & \textbf{$(R,D)$} & \textbf{$(D,L)$} & \textbf{$(D,R)$} & \textbf {Mean}\\
\hline
$\texttt{KG-only}$ & {56.0} & {27.3} & {40.5} & {28.1} & {39.5} & {56.2} & {41.3} \\
\hline
$\texttt{ARNN-GRU}^{*}$ & {52.9 (1.8)} & {40.4 (0.7)} & {40.4 (1.0)} & {35.1 (0.6)} & {51.1 (1.7)} & {48.4 (1.1)} & {44.7 (1.1)} \\
\hline
$\texttt{BERT}$ & {45.1 (3.6)} & {42.2 (0.5)} & {44.6 (1.9)} & {38.1 (1.3)} & {47.0 (2.2)} & {51.9 (2.2)} & {44.8 (2.0)}\\
\hline
$\texttt{TRNN-GRU}^{*}$ & {53.8 (0.9)} & {41.2 (1.1)} & {40.2 (0.8)} & {37.3 (0.9)} & {51.7 (1.3)} & {51.2 (1.0)} & {45.9 (1.0)} \\
\hline
$\texttt{DeBERTa}$ & {54.3 (1.7)} & {40.5 (1.4)} & {47.5 (2.3)} & {39.6 (1.6)} & {47.1 (2.1)} & {54.5 (2.2)} & {47.3 (1.9)} \\
\hline
$\texttt{SA-EXAL}^{*}$ & {54.7 (2.0)} & {42.2 (0.5)} & {47.6 (1.9)} & {\textbf{40.5} (1.1)} & {47.7 (2.8)} & {54.5 (1.9)} & {47.9 (1.7)}\\
\hline
$\textbf{\texttt{DeBERTa-MA}}$ & {61.5 (1.4)} & {40.2 (1.1)} & {43.4 (2.5)} & {38.0 (1.8)} & {47.2 (1.1)} & {62.0 (0.7)} & {48.7 (1.4)} \\
\hline
$\textbf{\texttt{DeBERTa-PT}}$ & {66.0 (1.8)} & {41.0 (1.2)} & {49.7 (1.3)} & {38.5 (0.8)} & {52.5 (1.6)} & {64.9 (0.8)} & {52.1 (1.3)} \\
\hline
$\textbf{\texttt{BERT-PT}}$ & {\textbf{66.4} (1.1)} & {\textbf{42.3} (1.1)} & {\textbf{49.9} (1.4)} & {39.5 (1.8)} & {\textbf{55.3} (1.4)} & {\textbf{65.8} (0.7)} & {\textbf{53.2} (1.3)}\\
\hline
\end{tabular}}
\caption{Comparison of average aspect extraction F1 scores (with standard deviation in parentheses). An asterisk indicates previously-reported model results.}
\label{experiments:ki}
\end{table}

\begin{figure*}[h!]
     \centering
     \begin{subfigure}[b]{0.3\textwidth}
         \centering
         \includegraphics[width=\textwidth]{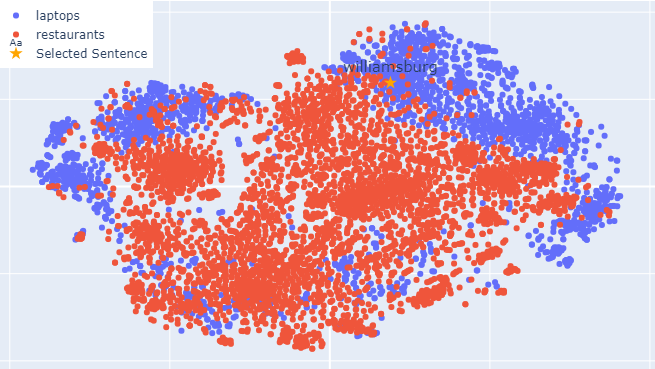}
         \caption{}
         \label{fig:bert_pt_emb}
     \end{subfigure}
     \hfill
     \begin{subfigure}[b]{0.3\textwidth}
         \centering
         \includegraphics[width=\textwidth]{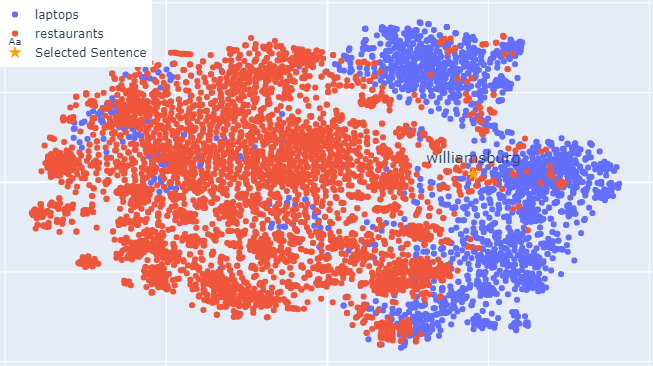}
         \caption{}
         \label{fig:bert_emb}
     \end{subfigure}
     \hfill
     \begin{subfigure}[b]{0.3\textwidth}
         \centering
         \includegraphics[width=0.85\textwidth]{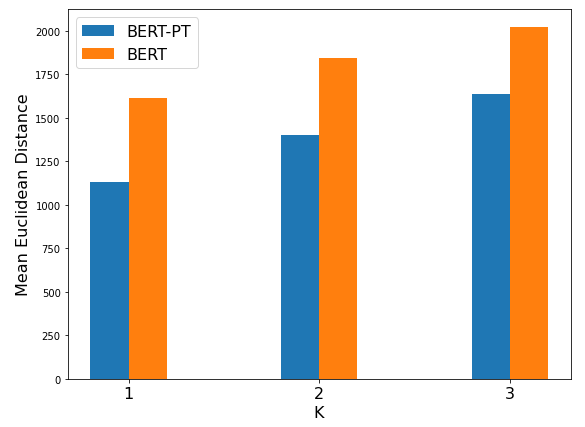}
         \caption{}
         \label{fig:k_closest}
     \end{subfigure}
        \cprotect\caption{Plots (a) and (b) depict t-SNE projections of the final hidden state embeddings of aspect terms produced by \verb|BERT-PT| and \verb|BERT| (respectively) in the laptops (purple) and restaurants (red) domains. Plot (c) shows the mean distance between aspects from one domain to the $K$-closest aspects in the other domain.}
        \label{fig:vis}
\end{figure*}

We evaluate the performance of state-of-the-art Transformer models on cross-domain aspect extraction when they are coupled with KGs using the two knowledge injection mechanisms detailed in Section~\ref{sec:injectionmechanisms}. The \verb|-PT| and \verb|-MA| suffixes within model names indicate the knowledge injection method used by our models, where \verb|-PT| denotes knowledge injection using pivot tokens and \verb|-MA| denotes knowledge injection via the modified attention scheme. We compare our knowledge-informed models to the following existing solutions for cross-domain ABSA:
\begin{itemize}
\itemsep0em 
    \item \verb|ARNN-GRU| \cite{Wang2018cross_domain}, a RNN architecture comprised of GRU blocks augmented with information from dependency trees through an auxiliary dependency relation classification task.
    \item \verb|TRNN-GRU| \cite{wang_pan_2019}, an extension of \verb|ARNN-GRU| incorporating a conditional domain adversarial network to explicitly align word feature spaces in the source and target domains.
    \item \verb|SA-EXAL| \cite{pereg-etal-2020-syntactically-libert}, a BERT-like model that incorporates syntactic information into its self-attention mechanism.
\end{itemize}
Additionally, we include several baseline models in our experiments as ablation studies. These include \verb|BERT| and \verb|DeBERTa| models that were fine-tuned on the ABSA task without knowledge injection as well as a \verb|KG-only| solution that classifies aspects solely based on our knowledge injection methodology. These ablations are discussed further in Section~\ref{sec:baselines}.

Table~\ref{experiments:ki} shows the mean and standard deviation of aspect extraction F1 scores in each cross-domain setting for our three knowledge-informed transformers and the baseline models. The best overall performance is obtained by \verb|BERT-PT|, which injects the knowledge by inserting pivot tokens into the input sequence. \verb|BERT-PT| achieves substantial improvements over the existing state-of-the-art \verb|SA-EXAL| with a $5\%$ absolute increase in mean F1 and a $10\%$ absolute F1 improvement when the restaurants dataset is the target domain.

\subsection{Analysis \& Discussion}

All three of our KG-enhanced models (\verb|BERT-PT|, \verb|DeBERTa-PT|, and \verb|DeBERTa-MA|) outperform the existing state-of-the-art solutions, which highlights the usefulness of our knowledge injection methods for improving aspect identification.
\verb|BERT-PT| performed better overall than both \verb|DeBERTa-PT| and \verb|DeBERTa-MA|, which is surprising because the baseline \verb|DeBERTa| model significantly outperformed the baseline \verb|BERT| model. 
These results suggest that the optimal knowledge-injection mechanism is highly model-dependent. 

One possible explanation for the lower performance of \verb|DeBERTa-PT| relative to \verb|BERT-PT| is that inserting tokens into the input sequence acts as a large disruption to the relative positional information used by the model. Inserting a pivot token shifts the relative positional encodings of every pair of tokens on opposite sides of it by one position. This can be contrasted with the absolute positional encodings used by BERT, for which the insertion of a pivot token affects only the positional encodings of the tokens following the pivot token. Moreover, because BERT's positional information is combined with the content information by summing the corresponding embeddings, we hypothesize that the inherent noisiness of this process would make the token representations more robust to changes in the positional encodings. 

The overall lower performance of \verb|DeBERTa-MA| relative to our pivot token models could be due to the knowledge-injection method introducing complexity overhead. The goal of this mechanism is to condition the attention by injecting a binary indicator for candidate aspect terms into the model. However, this requires learning high-dimensional embeddings and projection matrices to represent the binary indicator in the attention computation. The information conveyed by the binary indicators we use to identify candidate aspects may not be enough to offset the added model complexity in a way that improves upon the performance of using a pivot token. However, this mechanism could be beneficial when injecting more fine-grained and semantically-rich knowledge which can make use of the high-dimensional embedding space. We leave the exploration of this topic to future work.

\subsubsection{Impact of Knowledge Injection by Domain}

Knowledge injection yielded the greatest improvements in aspect identification when the restaurants dataset was used as the target domain. Differences in performance across domains can be attributed to variations in the size of the domain-specific KGs, the cardinality of the set of labeled aspects, and the consistency with which aspect tokens are labeled as aspects. Table~\ref{tab:kg_size_by_domain} provides a comparison of these three metrics across each domain. Aspect cardinality was measured as the number of unique aspect tokens occurring in the domain's test dataset. For each unique aspect, we measured the proportion of times the token was labeled as an aspect and averaged these proportions across all aspects to obtain the aspect consistency. The superior performance of \texttt{BERT-PT} in the restaurants domain appears to be driven by its much larger KG size relative to the other domains, which can be attributed to better coverage of food-related concepts in ConceptNet. The lower consistency of aspect usage in the devices domain is a major factor contributing to worse aspect extraction in this domain. We describe potential annotation differences between domains which may have caused this variation in aspect consistency in Section~\ref{sec:devices_data}.

\begin{table}[h!]
\centering
\resizebox{0.45\textwidth}{!}{%
\begin{tabular}{p{1.45cm} p{1.0cm} p{1.75cm} p{1.8cm}}
\hline
\textbf{Domain} & \textbf{KG Size} & \textbf{Aspect cardinality} & \textbf{Aspect consistency}\\
\hline
Restaurants & 8,547 & 1,321 & 0.78\\
Laptops & 4,651 & 847 & 0.77 \\
Devices & 5,824 & 516 & 0.49 \\
\hline
\end{tabular}
}
\caption{KG size, aspect set cardinality, and mean aspect consistency by target domain}
\label{tab:kg_size_by_domain}
\end{table}

\subsubsection{Transformer and KG Baselines}
\label{sec:baselines}

To explore the relative contributions of information embedded in the parameters of the Transformer model and the KG, we independently measure their performance at cross-domain aspect extraction. Specifically, we train baseline \verb|BERT| and \verb|DeBERTa| models on the same dataset described previously but without injecting knowledge. We also measure the performance of a \verb|KG-only| model that utilizes only the information conveyed by our knowledge injection methodology. In the \verb|KG-only| model, every token that is followed by a pivot token in our knowledge-enriched queries is labeled as an aspect as opposed to using a Transformer model to classify labels.

The results in Table~\ref{experiments:ki} show that \verb|DeBERTa| outperforms \verb|BERT| in five out of the six cross-domain settings. The \verb|KG-only| model performs best when the restaurants dataset is used as the target domain, even outperforming the other baseline models. These cross-domain results are consistent with those of our knowledge-informed Transformers, which also exhibit the greatest performance when the restaurants dataset is the target domain.

To illustrate the impact of external knowledge injection, Figure~\ref{fig:bert_pt_emb} and \ref{fig:bert_emb} depict the final hidden state embeddings of restaurant and laptop aspects produced by \verb|BERT-PT| and \verb|BERT| (respectively) after being projected to 2-D using t-SNE \cite{tsne-vandermaaten08a}. These visualizations show that \verb|BERT-PT| can better bridge the gap between aspects from the two domains, as evidenced by the increased overlap between representations of source and target domain aspects. This same effect is measured quantitatively in Figure~\ref{fig:k_closest}, which provides the mean Euclidean distance between the embedding of each aspect term and its closest $K \in \{1,2,3\}$ embeddings of aspects from the opposite domain.

\subsubsection{Stochastic Insertion of Pivot Information During Training}

As described in Section~\ref{sec:pivotphrase}, one component of our methodology involves stochastic insertion of the pivot token into the training dataset. 
Our motivation for using stochastic insertion is to adapt the model to differences in the accuracy and coverage of the KG in different domains. As shown in Table~\ref{experiments:ki}, the performance of the KG differs substantially across each of the three domains. This can be attributed both to differences in the coverage of the domain-specific KGs and the degree to which domain-specific words are labeled as aspects in each of the three domains, as described previously. 

An alternative to stochastic insertion of the pivot token into the training dataset is to assume both the training and testing domains have similar performance with respect to knowledge insertion. Under this assumption, the pivot token is inserted deterministically into the training dataset using the criteria described previously in Section~\ref{sec:whentoinject}. We conducted ablation studies on the training data insertion method and provide a detailed comparison of the aspect identification performance of our models under stochastic and deterministic pivot insertions in Table~\ref{tab:stochasticVsDeterministic}. Method $S$ corresponds to stochastic insertion of the pivot token during training while $D$ corresponds to deterministic insertion during training. These results show that stochastic insertion provides the greatest improvement in performance when the training domain KG performs worse than the testing domain KG. When the converse is true, slightly better performance is achieved by the deterministic insertion method.

\begin{table}[h!]
\centering
\resizebox{0.49\textwidth}{!}{%
\begin{tabular}{p{2.1cm} p{1cm} p{0.75cm} p{0.75cm} p{0.75cm} p{0.75cm} p{0.75cm} p{0.75cm}}
\hline
\textbf{Model} & \textbf{Method} & \textbf{$(L,R)$} & \textbf{$(L,D)$} & \textbf{$(R,L)$} & \textbf{$(R,D)$} & \textbf{$(D,L)$} & \textbf{$(D,R)$}\\
\hline
$\texttt{BERT-PT}$ & \centering \textit{S} & {\textbf{66.4}} & {42.3} & {49.9} & {39.5} & {\textbf{55.3}} & {\textbf{65.8}}\\
$\texttt{BERT-PT}$ & \centering \textit{D} & {47.3} & {\textbf{44.0}} & {\textbf{52.5}} & {41.8} & {48.8} & {52.8}\\
$\texttt{DeBERTa-PT}$ & \centering \textit{S} & {66.0} & {41.0} & {49.7} & {38.5} & {52.5} & {64.9}\\
$\texttt{DeBERTa-PT}$ & \centering \textit{D} & {57.9} & {43.5} & {51.8} & {\textbf{43.2}} & {50.1} & {56.8}\\
$\texttt{DeBERTa-MA}$ & \centering \textit{S} & {61.5} & {40.2} & {43.4} & {38.0} & {47.2} & {62.0}\\
$\texttt{DeBERTa-MA}$ & \centering \textit{D} & {51.6} & {41.7} & {44.1} & {38.5} & {47.1} & {55.4}\\
\hline
\end{tabular}
}
\caption{Comparison of the average aspect extraction F1 score for our knowledge-informed Transformer models trained with stochastic (\textit{S}) and deterministic (\textit{D}) training set injection methods}
\label{tab:stochasticVsDeterministic}
\end{table}

\subsubsection{Improving Aspect Label Distribution in the Digital Devices Dataset}
\label{sec:devices_data}
Table~\ref{experiments:ki} show that all evaluated models perform the worst when the digital devices dataset is used as the target domain. We believe this is partially attributable to a difference in the distribution of labels in the devices domain relative to that of the restaurant and laptop domains. Specifically, we observe that only 37\% of instances within the devices dataset contain a labeled aspect, whereas 50\% of instances in the laptop domain and 66\% of instances in the restaurant domain contain aspect labels. 

One possible cause for this inconsistency is differences in the annotation process used to collect the datasets. While the restaurant and laptop datasets were annotated under the same guidelines, the digital device reviews were collected nearly ten years earlier using different annotation instructions. During annotation of the devices dataset, aspects were only labeled for sentences in which the writer expresses an opinion \cite{hu2004mining}. This requirement was not specified in the annotation guidelines for the other two domains \cite{pontiki-etal-2014-semeval}, which may explain why the devices dataset has fewer aspect labels.

Motivated by this observation, we asked crowdsourced workers from Amazon Mechanical Turk to label the devices dataset in an effort to identify missing aspects. Each sentence in the dataset was labeled by five workers. New aspect labels were created when there was a majority agreement among the workers, which were then used to supplement the set of labels in the original devices dataset. We provide additional details on the collection of these new annotations in the appendix.

As a result of this process, the percentage of device reviews containing an aspect increased from 37\% to 57\%. Table~\ref{tab:newdevices} compares our knowledge-informed models to baseline Transformer models on cross-domain aspect extraction tasks utilizing this improved devices dataset (denoted $D'$). All methods performed better overall, indicating that the increased labeling consistency across the datasets improves domain transfer. We believe that our release of this updated devices dataset will facilitate future ABSA research by reducing annotation inconsistencies between domains. 

\begin{table}[t]
\centering
\resizebox{0.5\textwidth}{!}{%
\begin{tabular}{p{2.1cm} p{0.8cm} p{0.8cm} p{0.8cm} p{0.8cm} p{0.8cm} p{0.8cm} p{0.8cm}}
\hline
\textbf{Model} & \textbf{$(L,R)$} & \textbf{$(L,D')$} & \textbf{$(R,L)$}  & \textbf{$(R,D')$} & \textbf{$(D',L)$} & \textbf{$(D',R)$} & \textbf {Mean}\\
\hline
$\texttt{BERT}$ & {45.1 (3.6)} & {53.1 (0.7)} & {44.6 (1.9)} & {44.3 (2.0)} & {\textbf{59.1} (1.8)} & {57.2 (2.6)} & {50.5 (4.2)}\\
\hline
$\texttt{DeBERTa}$ & {54.3 (1.7)} & {52.3 (0.3)} & {47.5 (2.3)} & {44.3 (2.7)} & {57.1 (1.4)}  & {60.9 (2.5)} & {52.7 (1.6)}\\
\hline
$\texttt{DeBERTa-MA}$ & {61.5 (1.4)} & {49.9 (0.7)} & {43.4, (2.5)} & {41.4 (2.4)} & {50.1 (1.4)} & {63.9 (1.0)} & {51.7 (1.6)}\\
\hline
$\texttt{DeBERTa-PT}$ & {66.0 (1.8)} & {53.8 (1.0)} & {49.7 (1.3)} & {46.5 (1.3)} & {53.6 (1.9)} & {66.0 (1.0)} & {55.9 (1.4)}\\
\hline
$\texttt{BERT-PT}$ & {\textbf{66.4} (1.1)} & {\textbf{56.1} (1.0)} & {\textbf{49.9} (1.4)} & {\textbf{46.7} (1.1)} & {56.5 (2.1)} & {\textbf{66.8} (1.1)} & {\textbf{57.0} (1.3)} \\
\hline
\end{tabular}}
\caption{Comparison of average aspect extraction F1 scores (with standard deviation in parentheses) using improved devices dataset ($D'$).}
\label{tab:newdevices}
\end{table}

\subsubsection{Impact of Model and External Knowledge Size}
\label{sec:kgAndModelSize}

A recent trend in language modeling is the use of increasingly larger Transformers to achieve state-of-the-art performance on benchmark datasets, which has raised questions about the sustainability of this continued growth in model size. To explore the potential for external knowledge sources to mitigate this trend, we investigate the effect that increasing the size of the knowledge graph has on aspect extraction performance and contrast this with an increase in the model size of the Transformer. 

\begin{figure}[h!]
     \centering
         \includegraphics[width=0.45\textwidth]{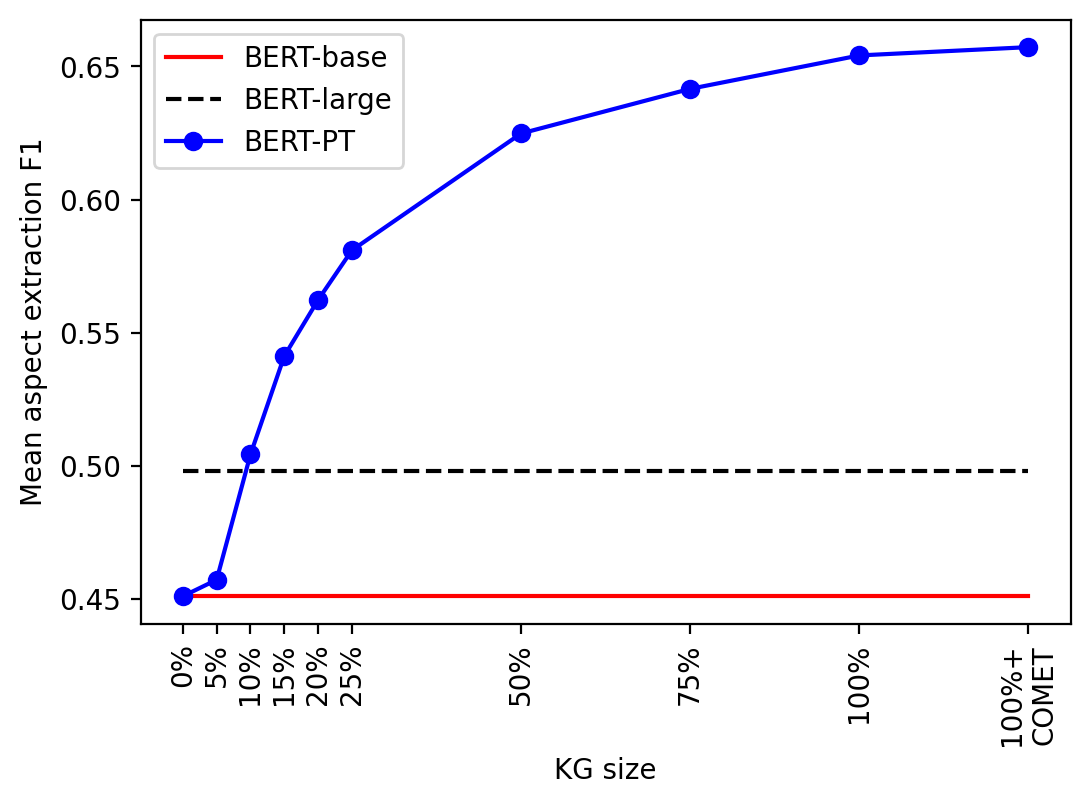}
         \caption{Target domain aspect extraction F1 in the $(L,R)$ experimental setting for different KG and Transformer sizes.}
         \label{fig:aspect_f1_kg_model_size}
\end{figure}

Figure ~\ref{fig:aspect_f1_kg_model_size} shows the results of an ablation study on KG and Transformer size using the $(L,R)$ experimental setting, in which the laptops dataset is used as the source domain and the restaurants dataset is used as the target domain. The target domain aspect extraction F1 of our \texttt{BERT-PT} model is provided across varying sizes of the knowledge graph used to inject the pivot token. KG sizes less than 100\% were obtained by randomly sampling a percentage of the original target domain KG obtained from ConceptNet. The 100\%+COMET result corresponds to using the full target domain KG, which includes all triples extracted from ConceptNet as well as automatically-generated triples produced by COMET. Note that we only vary the size of the KG used to inject knowledge into the target domain test dataset in order to simulate the effect of altering the knowledge source without retraining the model.

We observe that the aspect extraction F1 of \texttt{BERT-PT} monotonically increases with the size of the KG, which suggests that expanding the amount of external knowledge available to the model can lead to improved performance at inference time without requiring the Transformer to be retrained. In contrast, increasing the size of a \texttt{BERT} model which does not leverage external knowledge from 110M parameters (\texttt{BERT-base}) to 336M parameters (\texttt{BERT-large)} produces a much smaller improvement in aspect extraction F1. Our \texttt{BERT-PT} model outperforms \texttt{BERT-large} when as little as 10\% of the original ConceptNet subgraph is used as the KG, despite \texttt{BERT-large} having over 3 times more parameters than \texttt{BERT-PT}.

\begin{figure*}[h!]
     \centering
     \begin{subfigure}[b]{0.49\textwidth}
         \centering
         \includegraphics[width=\textwidth]{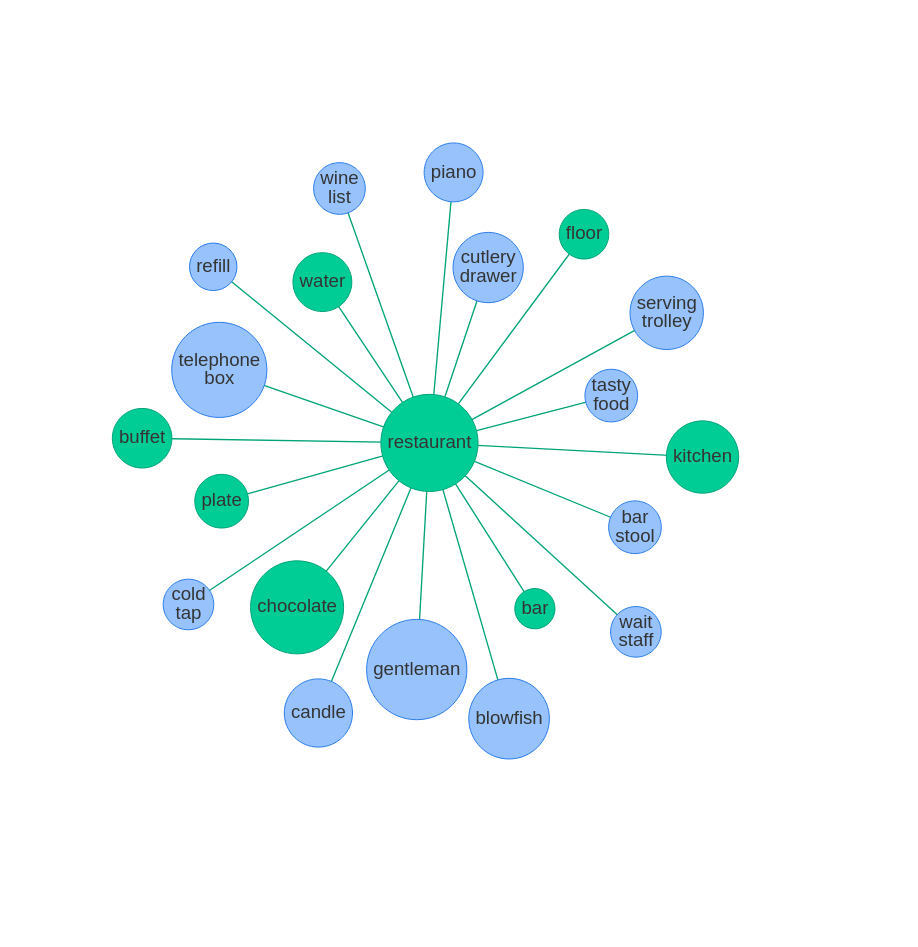}
         \caption{}
     \end{subfigure}
     \hfill
     \begin{subfigure}[b]{0.49\textwidth}
         \centering
         \includegraphics[width=\textwidth]{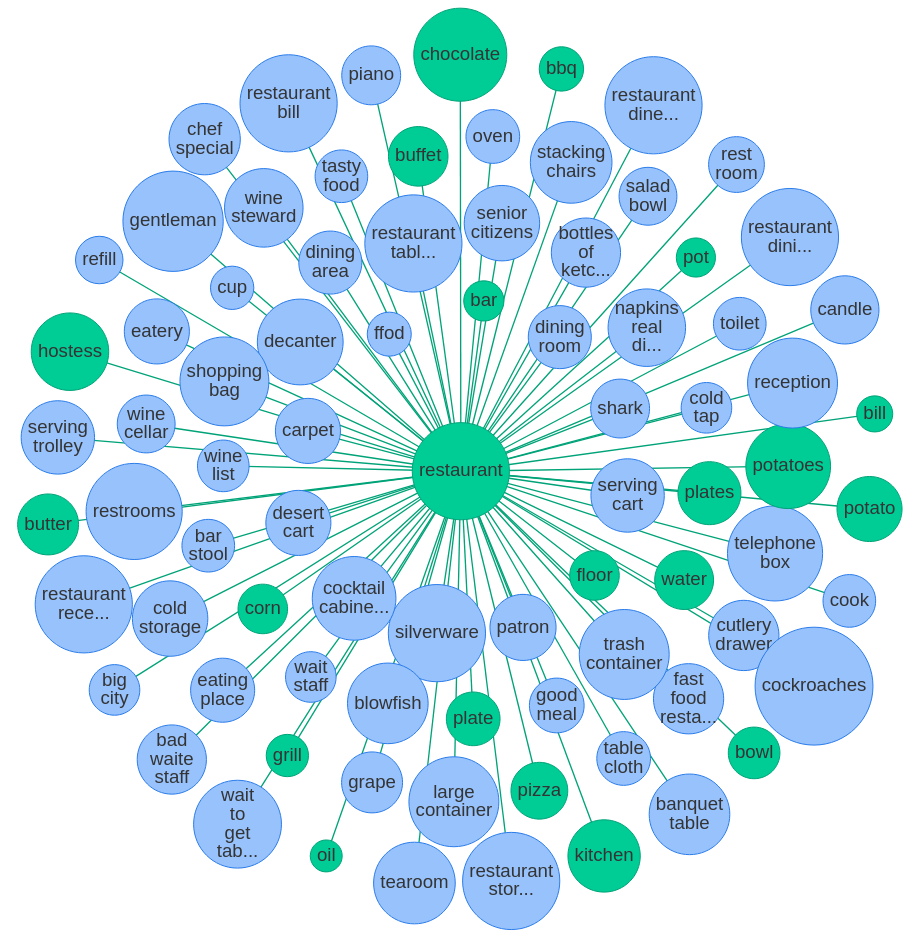}
         \caption{}
     \end{subfigure}
     \hfill
     \cprotect\caption{Visualizations of a 25\% sampled subgraph (a) and the full subgraph (b) of ConceptNet nodes connected to the "restaurant" concept by the "AtLocation" relation. Green nodes correspond to labeled aspect tokens in the restaurants domain while non-aspect tokens are represented by blue nodes.}
        \label{fig:kgLaptopsToRestaurants}
\end{figure*}

To illustrate the difference in size of KGs used in this analysis, Figure~\ref{fig:kgLaptopsToRestaurants} visualizes the 25\% and 100\% KGs produced for the restaurant domain. Due to the large size of the full KGs, only a subgraph consisting of triples matching the pattern $(x, AtLocation, restaurant)$ are depicted. Green nodes correspond to labeled aspect tokens in the restaurants domain while non-aspect tokens are represented by blue nodes. The 25\% KG lacks many nodes that correspond to labeled aspects, including tokens such as "hostess", "butter", and "pizza". However, many of these missing nodes could be easily added by human annotators in order to improve the performance of the model without the need for retraining. Such human-in-the-loop systems are a promising research direction for future studies on Transformers augmented with external knowledge sources. 

\subsubsection{Analysis of errors in the $(L,R)$ experimental setting}

In our experiments, \texttt{BERT-PT} achieves its greatest performance when the restaurants dataset is used as the target domain. To better understand how the use of external knowledge is improving aspect extraction in this domain, we investigated which aspects are correctly identified or missed by \texttt{BERT} and \texttt{BERT-PT} in the $(L,R)$ setting.

Table~\ref{tab:lr_tp_and_fp} provides the top-10 most frequent true positive (TP) and false negative (FN) aspect tokens identified by \texttt{BERT} and \texttt{BERT-PT}. In this analysis, a TP is defined as a token which is labeled as either aspect type and is correctly predicted to be an aspect. A FN is defined as tokens which are labeled as an aspect, but are not predicted to be an aspect. Counts of the number of TP and FN occurences for each token are expressed in parentheses in Table~\ref{tab:lr_tp_and_fp}.

\begin{table}[t]
\centering
\resizebox{0.45\textwidth}{!}{%
\begin{tabular}{|p{2.25cm}|p{2.1cm}|p{2.25cm}|p{2.1cm}|}
\hline
\multicolumn{2}{|c|}{\textbf{\texttt{BERT}}} & \multicolumn{2}{|c|}{\textbf{\texttt{BERT-PT}}} \\
\hline
\multicolumn{1}{|c|}{\texttt{TP}} & \multicolumn{1}{|c|}{\texttt{FN}} & \multicolumn{1}{|c|}{\texttt{TP}} & \multicolumn{1}{|c|}{\texttt{FN}} \\
\hline
service (1066) & food (1288) & food (1797) & place (325) \\
food (527) & place (376) & service (1075) & of (196) \\
staff (318) & of (202) & staff (330) & with (141) \\
menu (189) & pizza (196) & menu (239) & and (138) \\
atmosphere (145) & dinner (174) & pizza (216) & food (129) \\
prices (135) & dishes (165) & atmosphere (215) & dinner (122) \\
price (114) & wine (153) & wine (213) & the (114) \\
decor (110) & with (147) & dishes (156) & indian (106) \\ 
wine (96) & chicken (135) & prices (153) & lunch (93) \\
list (74) & and (135) & sushi (134) & dumplings (76) \\
\hline
\end{tabular}}
\caption{The top-10 true positive (TP) and false positive (FP) aspects identified by \texttt{BERT} and \texttt{BERT-PT} in the $(L,R)$ test dataset. Frequency counts are provided in parentheses.}
\label{tab:lr_tp_and_fp}
\end{table}

Both \texttt{BERT} and \texttt{BERT-PT} share many of the same TP aspect tokens. However, \texttt{BERT-PT} correctly classifies more of these aspects than \texttt{BERT}, which suggests that the injection of knowledge helps improve the consistency with which the model correctly classifies aspects in the target domain. A higher proportion of the TP aspects identified by \texttt{BERT} are not specific to the target domain (e.g., \lq service', \lq price'), whereas \texttt{BERT-PT} correctly identifies more domain-specific aspects (e.g., \lq pizza', \lq sushi'). Finally, \texttt{BERT-PT} has a higher proportion of stopwords among its top FN aspects, suggesting that more of its errors are associated with tokens that are infrequently labeled as aspects as opposed to the domain-specific aspect words more frequently missed by \texttt{BERT}.

\begin{figure}[h!]
     \centering
         \includegraphics[width=0.45\textwidth]{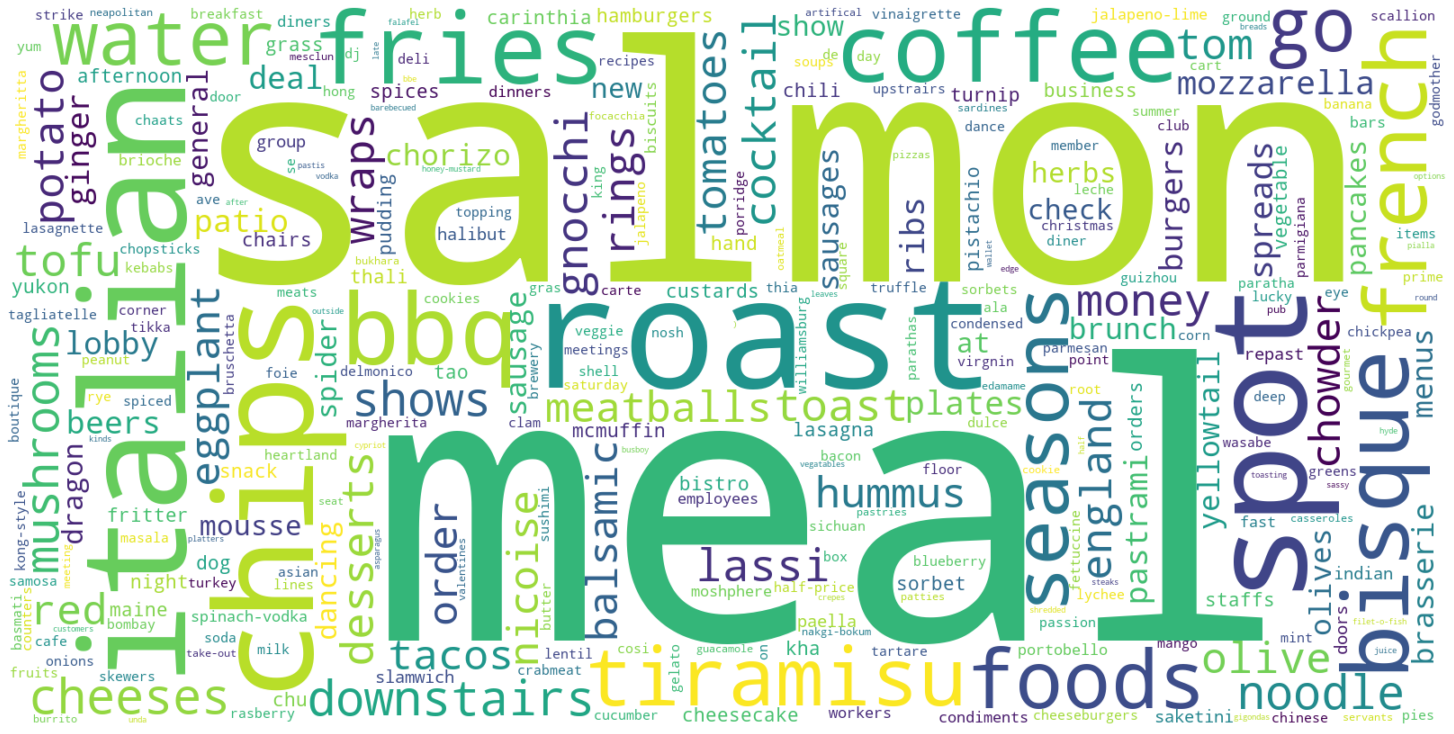}
         \caption{Wordcloud visualization of our most common newly-identified aspects in the restaurants domain}
         \label{fig:lr_new_aspects_wordcloud.png}
\end{figure}

Many of the aspects identified by \texttt{BERT-PT} are never labeled as aspects by \texttt{BERT}. Figure~\ref{fig:lr_new_aspects_wordcloud.png} provides a wordcloud visualization of such newly-identified aspects by our model. The size of each aspect token depicted in the wordcloud is proportional to its frequency of occurrence in the restaurants test dataset. This figure shows that \texttt{BERT-PT} identifies a broad range of new aspect words specific to the restaurants domain, which can be attributed to the breadth of knowledge made available to the model through the KG.

\section{Conclusion}

We have presented a comprehensive approach for constructing domain-specific KGs and determining when it is beneficial to inject knowledge into Transformers for aspect extraction. Additionally, we introduced two alternative approaches for knowledge injection: via query enrichment and using a disentangled attention mechanism. Our experimental results show that our knowledge-informed Transformers outperform existing state-of-the-art models on cross-domain aspect extraction tasks. Finally, we release an improved version of the benchmark digital device reviews dataset to support future research on aspect-based sentiment analysis.

While this work focused on identifying aspects for ABSA, our domain-specific KG construction and knowledge injection methods can be applied to other NLP tasks for which external knowledge is beneficial. In future work, we intend to explore extensions of our approach to such applications and investigate alternative methods for injecting knowledge into language models.

\appendix

\section{Appendix}
\label{sec:appendix}

\subsection{Crowdsourcing Annotations for Improved Devices Dataset}

Annotations for the digital devices dataset were collected from crowdsourced workers using Amazon Mechanical Turk. Workers were paid a reward of \$0.14 per annotated sentence and were required to meet certain criteria in order to work on the task. Specifically, workers were required to have a Human Intelligence Task (HIT) approval rate > 95\%, completed at least 5000 HITs, and obtained the Mechanical Turk Masters qualification.

Workers were asked to apply aspect and opinion labels by highlighting portions of each sentence. Aspect labels were further segmented into positive, negative, neutral, and conflict polarities in accordance with prior labeling tasks for ABSA. Detailed instructions were provided on how to identify aspect and opinion words, including multiple examples from the laptop reviews dataset. 

Overlapping text selections were required from at least 3 out of the 5 workers to generate a new label. For multi-word phrases, we took the longest subset of the phrase which was selected by at least 3 workers as the label. New aspect annotations were added to the existing set of aspect labels to produce the improved devices dataset. In cases where multi-word labels overlapped between the original and new label set, we kept the longer form of the label. 

\bibliographystyle{ACM-Reference-Format}
\balance
\bibliography{custom}

\end{document}